\documentclass[letterpaper, 10 pt, conference]{ieeeconf}  

\IEEEoverridecommandlockouts                              

\usepackage{graphics} 
\usepackage{epsfig} 
\usepackage{mathptmx} 
\usepackage{times} 
\usepackage{amsmath} 
\usepackage{amssymb}  

\usepackage{algorithm}
\usepackage{algorithmic}

\usepackage{tabularx}

\title{\LARGE \bf
MMRDN: Consistent Representation for Multi-View Manipulation Relationship Detection in Object-Stacked Scenes
}

\author{Han Wang$^{1}$, Jiayuan Zhang$^{1}$, Lipeng Wan$^{1}$, Xingyu Chen$^{1}$, Xuguang Lan$^{1}$, Nanning Zheng$^{1}$
\thanks{$^{1}$Xuguang Lan is with National Key Laboratory of Human-Machine Hybrid Augmented Intelligence, National Engineering Research Center for Visual Information and Application, Institute of Artificial Intelligence and Robotics Xi'an Jiaotong University, Xi'an, Shaanxi, China
        {\tt\small xglan@mail.xjtu.edu.cn}}%
}

\begin{document}

\maketitle
\thispagestyle{empty}
\pagestyle{empty}

\begin{abstract}

Manipulation relationship detection (MRD) aims to guide the robot to grasp objects in the right order, which is important to ensure the safety and reliability of grasping in object stacked scenes. 
Previous works infer manipulation relationship by deep neural network trained with data collected from a predefined view, which has limitation in visual dislocation in unstructured environments. Multi-view data provide more comprehensive information in space, while a challenge of multi-view MRD is domain shift. In this paper, we propose a novel multi-view fusion framework, namely multi-view MRD network (MMRDN), which is trained by 2D and 3D multi-view data. We project the 2D data from different views into a common hidden space and fit the embeddings with a set of Von-Mises-Fisher distributions to learn the consistent representations. Besides, taking advantage of position information within the 3D data, we select a set of $K$ Maximum Vertical Neighbors (KMVN) points from the point cloud of each object pair, which encodes the relative position of these two objects. Finally, the features of multi-view 2D and 3D data are concatenated to predict the pairwise relationship of objects. Experimental results on the challenging REGRAD dataset show that MMRDN outperforms the state-of-the-art methods in multi-view MRD tasks. The results also demonstrate that our model trained by synthetic data is capable to transfer to real-world scenarios.

\end{abstract}

\let\thefootnote\relax\footnotetext{This work was supported in part by National Key R\&D Program of China under grant No. 2021ZD0112700, NSFC under grant No.62125305, No.62088102, and No.61973246, the Fundamental Research Funds for the Central Universities under Grant xtr072022001.}


\section{INTRODUCTION}

\begin{figure}
  \centering
  \includegraphics[width=0.47\textwidth]{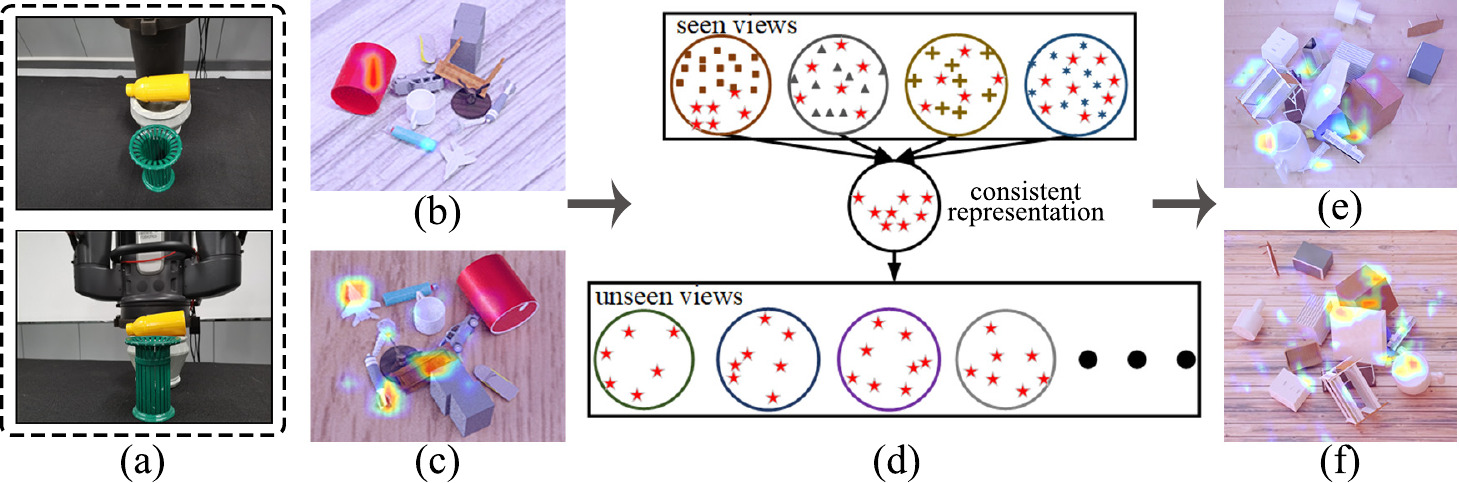}
  \caption{(a) Visual dislocation. (b) and (c)The feature GRAD-CAM heatmaps of the model trained by VMRN\cite{vmrn} in the same scene from different views. (d) The fundamental of our method. (e) and (f) the feature GRAD-CAM heatmaps of our model in the same scene from different views.}
  \label{vis_dis}
\end{figure}

Grasping is a basic and crucial skill for various robot manipulation tasks. In object stacked scenes, the robot is required to infer a proper grasping order for safe and reliable manipulation, which introduces the problem of manipulation relationship detection (MRD). 
Currently, the visual detection of the grasp region in unstructured environment attracts many efforts \cite{background}. In unstructured environment, the visual dislocation like the phenomenon shown in Fig.\ref{vis_dis}(a) may occur, which results in erroneous detection of MRD.
Therefore, it is significant to guarantee the correct identification of MRD when visual dislocation is inevitable.

Recently, deep-learning-based methods have achieved great success on MRD in single-view grasping.
They encode the objects and object pairs in rgb images and fuse the high-dimensional features of them to infer manipulation relationships.
While, most deep learning-based MRD algorithms like VMRN\cite{vmrn} and GGNN-VMRN\cite{ggnn-vmrn} have limitations in MRD from different views. For example, Fig.\ref{vis_dis}(b) shows the heat maps of the VMRN\cite{vmrn}, indicating that the representation of manipulation relationships is different from different views. Multi-view data are able to alleviate the above problem for that more information can be obtained.
Yang et al.\cite{mvda} indicate that the view consistency based on source data (training data) is largely violated in the target domain (test data) due to the distribution gap between different domain data. Therefore, a challenge of multi-view MRD is the consistent representation learning of manipulation relationships between different domains. 

To address the problem above, we proposes a novel multi-view fusion framework to strengthen the views’ consistency by identifying the associations of domain-specific features related to manipulation relationships from different domains.
First, from the definition of the manipulation relationships\cite{vmrn}, it can be known that the position information of objects in space is crucial for MRD, so we define a set of $K$ Maximum Vertical angle Neighbors (KMVN) points from the point cloud to represent relative position of each object pair.
For the 2D data, we project the features of images and objects from different views into a common hidden space and fit the embedings with a set of Von-Mises-Fisher distributions. Such distributions are aligned to reduce the representation variance of data from different domains. Finally, the features of multi-view 2D and 3D data are concatenated to predict the pairwise relationship of objects.

In summary, we have two main contributions in this paper:
\begin{itemize}
\item We propose a novel framework to detect manipulation relationships from multi-view data in object stacked scenes. Our framework is proved effective to alleviate the problem of domain shift under different views.

\item Experimental results show that MMRDN achieves state-of-the-art performance on the REGRAD dataset (not only the data from seen views but the data from unseen views) and our model can be transferred into the real world data.
\end{itemize}

\section{Related Work}

\subsection{Visual manipulation relationship detection}
Visual manipulation relationship detection is proposed by \cite{vmrn} and aims to infer a proper grasping order for safe and reliable manipulation.
\cite{vmrn} learn the relative positions between objects through features of RGB images and constructed a manipulation relationship tree through images inputting. \cite{crf} added Condition Random Field in the process of relational reasoning. \cite{ggnn-vmrn} encode the global context information and position information of object pairs by transformer. Then Gated Graph Neural Network is applied to fuse the encoding features.
\cite{regrad} collect a large scale dataset in virtual environment for robotic grasping. They verify the VMRN\cite{vmrn} in multi-view data while they ignore the domains shift in multi-view data. Also, spatial relationship can be applied in MRD. \cite{sr} determines the relative position between object pairs by xyz coordinates of the 3D point cloud of the object in space. But the application scenarios of \cite{sr} are simple.

\subsection{Multi-source domain adaptation}
Domain adaptation assumes data comes from both a source domain and a target domain, but different distributions are hold in different domains. \cite{mmd} and \cite{dann} work well to solve the shift between source and target domains.
Multi-source domain adaptation (MSDA) considers a generalized case that models generalization ability as more diverse data included but more challenging since domain shift also exists among source domains. 
\cite{hoffman,mansour,sun,sun2013} handling this problem through a weighted source combination to achieve target-relevant prediction with rigorous theoretical analysis. \cite{peng} dynamically aligns moments of feature distributions, which consist of pairs of source and target domains and those of source domains. Rather than explicit feature alignment, \cite{venkat} uses pseudo-labeled target samples for implicit alignment.
DMSN\cite{DMSN} introduce MSDA into object detection. It develops feature alignment among sources and pseudo subnet learning for their weighted combination. TRKP\cite{tpkp} aims at preserving more target-relevant knowledge from different source domains to facilitate multi-source DAOD.

\subsection{Multi-view learning}
Multi-view learning like co-training mechanism\cite{co-train}, subspace learning methods\cite{clus}, and multiple kernel learning (MKL)\cite{kernel} as well as algorithm with deep learning aims to integrate multi-view information from different views so as to obtain more discriminative common representations. It has been applied in the domain of video surveillance, entertainment media, social networks and medical detection,  while it has not been in manipulation relationship detection.

\begin{figure*}
  \centering
  \includegraphics[width=1.\textwidth]{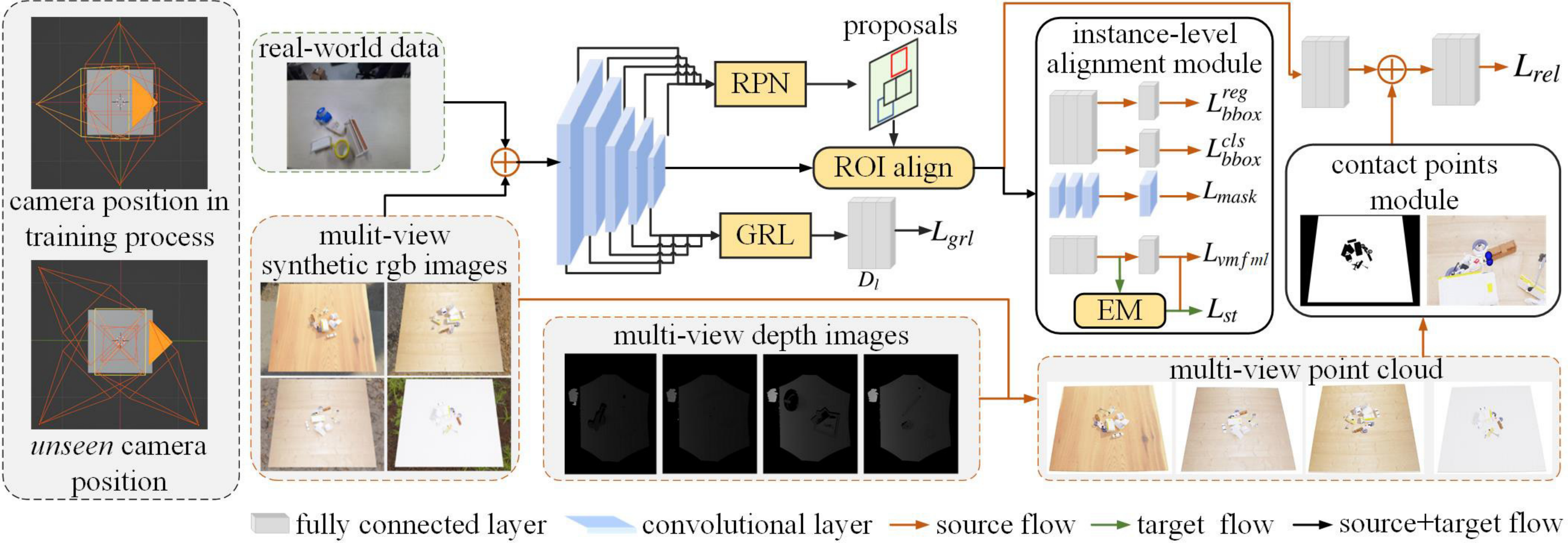}
  \caption{Architecture of Multi-View Manipulation Relationship network (MMRDN). Orange and green arrows indicate forward flows for source domain and target domain respectively. Black arrows indicates the forward flows of both source and target domains. Instance-level alignment module is designed to align image features and object features from different domains. Contact point module (CPM) is designed to select KMVN from point cloud to represent relative position of object pairs. The left dotted box shows the camera positions used during training process and the \textit{unseen} views that appeared in Table.\ref{tab1}}
  \label{Network}
\end{figure*}

\section{Method}

\subsection{Overview} \label{Overview}
In multi-view MRD, we consider the scenario that there are $N_s$ labelled source domains $S_1, S_2, \cdots, S_{N_s}$ and one unlabelled target domain $T$. In the $i$-th source domain $S_i=\{ (x_{r_i}^j, x_{d_i}^j, B_i^j, m_i^j, y_i^j, r_i^j) \}_{j=1}^{N_i}$, suppose $x_{r_i}^j, x_{d_i}^j, B_i^j, m_i^j, y_i^j, r_i^j$ are RGB images, depth images, bounding boxes, segmentation mask, category labels and relationship labels respectively. Note that $N_i$ is the number of images in the $i$-th source domain. In the unlabelled target domain $T=\{ x_{r_T}^j \}_{j=1}^{N_T}$, the $j$-th image is represented by $x_{r_T}^j$, and $N_T$ denotes the images of target domain. In this problem, our goal is to learn a manipulation relationship detector that can correctly detect the object and identify the manipulation relationship from an arbitrary view based on multiple labelled source domain and unlabelled target domain.

We propose a novel framework termed multi-view manipulation relationship detection network (MMRDN) and the pipeline is shown in Fig.\ref{Network}. It has the following distinct characteristics. For that labelled real-world data are collected costly, synthetic data containing RGB and depth images from four different views and real-world data containing RGB images from a single view are input into the network.
The consistent representation learning process is divided into three parts. In the first part, we align image-level features for all the domains. In the second part, instance-level features in source domains are aligned by VMFML\cite{vmfml} and the features from source domains and target domains are aligned by cosine similarity measure between source domains and target domain.In the third part, we construct the features of the relative position between object pairs in the contact point module. Finally, features from instance-level alignment module and contact point module are fed into the classifier to learn the manipulation relationships.

\subsection{Image-level Alignment Module} \label{grl}

We aim to learn domain-invariant features with shared parameters. Considering the fact that low-level (image-level) features are scarcely associated with high-level semantics, and low-level features benefit to improve the localization ability. Therefore, we conduct aligning local features in lower layers using a cross-entropy loss to train the domain discriminator $D_l$. A gradient reversal layer (GRL)\cite{dann} is placed between the backbone and the domain discriminator to implement adversarial learning. The cross-entropy loss is formulized as:
\begin{equation}
L_{grl} = -\sum_{i=1}^{N_l} \sum_{j=1}^{N_s+1} \sum_{k=1}^K y_{i,j,k} \log(p_{i,j,k})
\end{equation}
where $N_l$ denotes the number of layers of the final output features. $N_s$ and $K$ denote the number of source domains and classes, respectively. $N_s+1$ denotes the number of source domains and one target domain.

\subsection{Instance-level Alignment Module} \label{vmfml}

Spherical feature embedding retains the power of feature learning because it only reduces feature dimension by one but makes domain adaptation easier since differences in norms are eliminated. Therefore, we utilize the Von-Mises-Fisher (VMF) distribution, which is a unit spherical normal distribution, to align the instance-level features in different domains.

For that source samples are labelled and target samples are unlabelled, Von-Mises-Fisher Model Loss (VMFML) and EM algorithm are applied for source domains and target domain, respectively,  to fit the embeddings into VMF distribution. The bounding box classification gain knowledge of the target domain data via EM algorithm, but in the test or inference phase, we inference a scene though the trained classifier.

\subsubsection{Von-Mises-Fisher Model Loss}

A VMF distribution is defined as:
\begin{equation}
	p(\mathbf{z}|\mu,\kappa) = C_d (\kappa) \exp{(\kappa \mu^T \mathbf{z})} \label{p}
\end{equation}
\begin{equation}
	C_d (\kappa) = \frac{\kappa^{\frac{d}{2}-1}}{(2\pi)^{\frac{d}{2}} I_{\frac{d}{2}-1}(\kappa)} \label{cd}
\end{equation}
where  $||\mu||_2=1$ represents the mean direction on the unit sphere, $\kappa \in \mathbb{R}_{\geq 0}$ represents the concentration around $\mu$, and $I_v$ is the modified Bessel function of the first kind and order $v$. 

Then the VMF Mixture Model (VMFMM) with M classes is defined as \cite{vmfmm}:
\begin{equation}
    g_v(\mathbf{z}_i| \Theta_K) = \sum_{j=1}^K \pi_j p(\mathbf{z}_i|\mu_j, \kappa_j) \label{g_v}
\end{equation}
where $\Theta = \{ (\pi_1, \mu_1, \kappa_1), \cdots, (\pi_K, \mu_K, \kappa_K) \}$ is the set of parameters, $\pi_j$ is the mixing proportion of the $j^{th}$ class.

For fairness in each category, the concentration coefficient $\kappa$ of each category has been to the same. So, posterior probability based on cross entropy guided by the VMFMM can be rewritten as:
\begin{equation} \label{psij}
    p_s^{ij} = \frac{\exp{(\kappa_j \mathbf{\mu}_j^T \mathbf{z}_i)}}{\sum_{l=1}^K \exp{(\kappa_l \mathbf{\mu}_l^T \mathbf{z}_i)}}
\end{equation}
where $p_s^{ij}$ is the probability of the $i^{th}$ sample in synthetic data belongs to the $j^{th}$ classes. And $\mathbf{z}_i = \frac{h_i}{||h_i||}$, $h$ denotes the latent variable from fully-connect layer; $\mathbf{\mu}_j = \frac{\mathbf{\omega}_j}{||\mathbf{\omega}_j||}$, $\mathbf{\omega}_j$ denotes the softmax weight of $j^{th}$ class; $\kappa_j$ denotes the concentration parameter of $j^{th}$ class.

As written in VMFML\cite{vmfml}, the loss function is :
\begin{equation}
L_{vmfml} = -\sum_{i=1}^N \sum_{j=1}^K y_{ij} \log(p_s^{ij}) \label{L_s_vmf}
\end{equation} 
where $y_{ij}$ is the one-hot label of $i^{th}$ in $j^{th}$ class.

\subsubsection{EM algorithm of VMF distribution}
The objective of EM algorithm is to estimate model parameters such that the negative log-likelihood value, i.e. $-log(g(\mathbf{z}_i|\Theta_K))$ is minimized. The EM method estimates the posterior probability in the E-step as\cite{vmfmm}:
\begin{equation}\label{ptij}
    p_t^{ij} = \frac{\pi_j C_d(\kappa_j) \exp{(\kappa_j \mu_j^T \mathbf{z}_i)}}{\sum_{l=1}^K \pi_l C_d(\kappa_l) \exp{(\kappa_l \mu_l^T \mathbf{z}_i)}}
\end{equation}
and model parameters in the M-step as\cite{vmfmm}:
\begin{equation}
\begin{aligned}
    &\pi_j = \frac{1}{N} \sum_{i=1}^N p_{ij}, \ \ \hat{\mu}_j = \frac{\sum_{i=1}^N p_{ij} \mathbf{z}_i}{\sum_{i=1}^N p_{ij}}, \\
    &\overline{r} = \frac{||\hat{\mu}_j||}{N \pi_j}, \mu_j = \frac{\hat{\mu}_j}{||\hat{\mu}_j||}, \kappa_j = \frac{\overline{r}d-\overline{r}^3}{1-\overline{r}^2}
\end{aligned}
\end{equation}

In order not to make the estimated parameter gap between the source domain and target domain data too large, different from the traditional EM algorithm which initializes the distribution parameters by means of spherical clustering during initialization, we use the parameters trained from the source domain data to initialize.

Finally, we minimize the gap between source and target domains by cosine distance.
\begin{equation} \label{L_st}
L_{st} = \sum_{j=1}^K \frac{1}{1 + \cos(\mathbf{\mu}_s^j, \mathbf{\mu}_t^j)}
\end{equation}

\subsection{Contact Points Module} \label{CPM}

Manipulation relationship should be the relative position in the 3D space. Under the circumstance that the full point cloud and shape of the object are unavailable, KMVN is selected from the partially observable point cloud of objects to guide the network to focus on contact part for object pairs.

\textbf{$K$ Maximum Vertical Neighbors point set}  \label{cp-def}
Given two objects, $O_i=\{o_i^1, o_i^2, \cdots, o_i^n \}$ and $O_{i'} = \{o_{i'}^1, o_{i'}^2, \cdots, o_{i'}^m \}$, where $o_i^j= (x_i^j, y_i^j, z_i^j )$ is the coordination of the j-th point of the i-th object and another analogy. The vertical angle of all direction vectors between two objects can be denoted as $V_{ii'}^p = \pi - \arccos(O_i - O_{i'}, z)$ where $z$ denotes the vertical axis in the coordinate System and $p \in \mathcal{R}^{n\times m}$. if $\mathcal{O}_i^k \subseteq O_{i}$ and $\mathcal{O}_{i'}^k \subseteq O_{i'}$ satisfy 
\begin{align}
V_{ii'}^k &= top_k{V_{ii'}^p}  \\
V_{ii'}^k &= \pi - \arccos(\mathcal{O}_i^k - \mathcal{O}_{i'}^k, z)
\end{align}
Then $\mathcal{O}_i^k$ of object $i$ and $\mathcal{O}_{i'}^k$ of object $i'$ is called $K$ Maximum Vertical Neighbors point set.

\begin{figure}[htbp]
\centering
\includegraphics[width=0.47\textwidth]{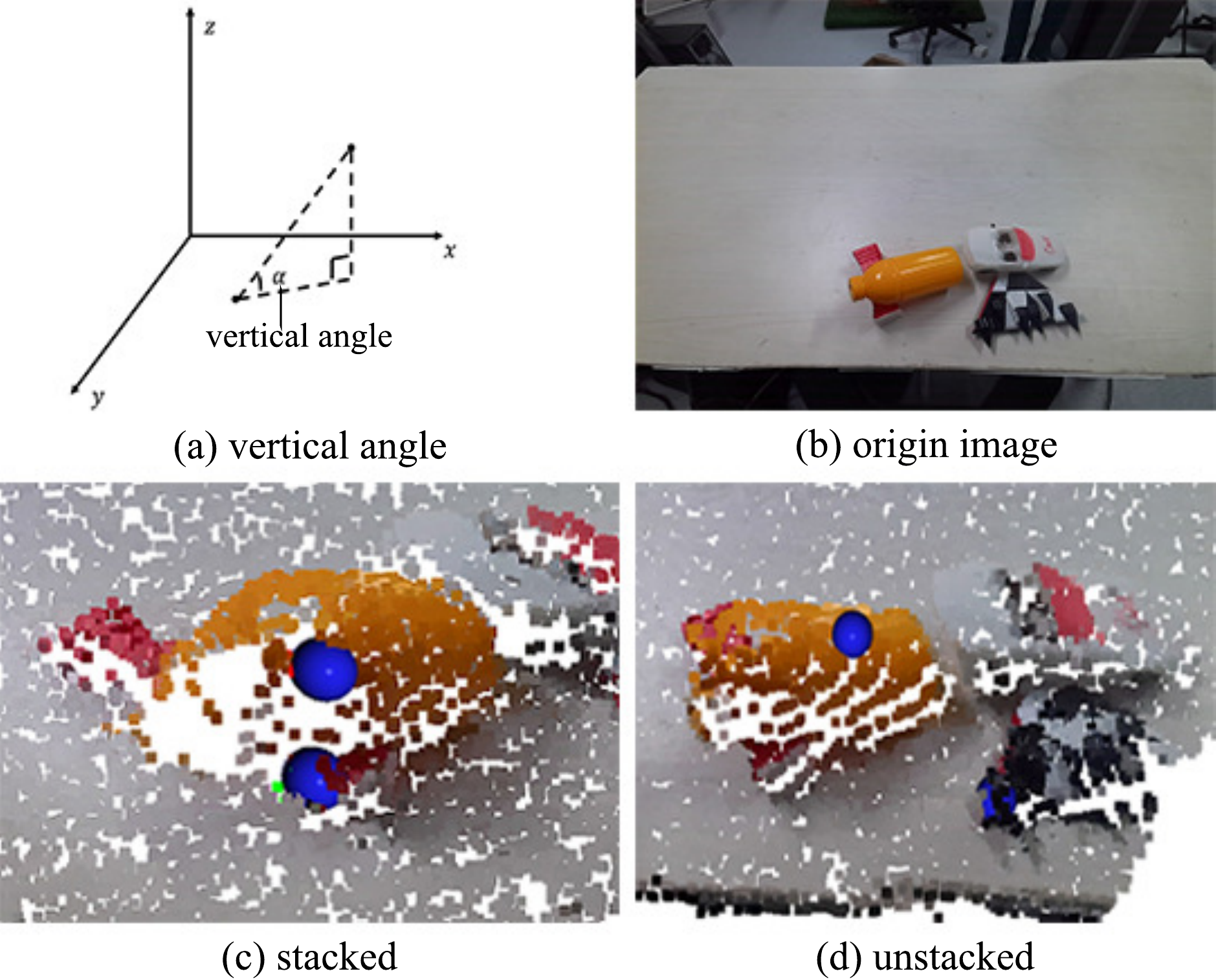}
\label{cps-a}
\caption{Explain of vertical angle and examples of KMVN. The blue points denote the KMVN in object pairs. (a) The explanation of vertical angle,i.e. the angle between the sight line and its horizontal line of sight. (b) Scene to show KMVN, where the ``bottle'' is above the ``mailbox'' and the ``car'' is above the ``airplane''. (c) KMVN between ``bottle'' and ``mailbox'' which are stacked. (d) KMVN between ``bottle'' and ``airplane'' which are unstacked.}
\label{cp}
\end{figure}

Based on the above analysis, KMVN is the top $k$ points with the maximum vertical angle between the direction vector and the z-axis between all points in the two objects. We take an example in Fig.\ref{cp}. If two objects are stacked, then their KMVN will be closer to ``up and down'', on the contrary, if two objects are gradually moving away, then the KMVN between them will be closer to ``left and right''. That is to say, KMVN can well represent the relative position of object pairs.

Then the relative position containing direction and distance of object pairs can be represented as follows. 
 
\begin{equation}
Z_{op_1} = \frac{1}{k} \sum_k \cos(\mathcal{O}_i^k - \mathcal{O}_{i'}^k, z)
\end{equation}
\begin{equation}
Z_{op_2} = \frac{1}{k} \sum_k ||\mathcal{O}_i^k - \mathcal{O}_{i'}^k||_2
\end{equation}
\begin{equation}
Z_{U} = Z_{op_1} \oplus Z_{op_2}
\end{equation}
where $Z_{op_1}$ and $Z_{op_1}$ denote the representations of object pairs, `$\oplus$' denotes the concatenate. $Z_{op_1}$ and $Z_{op_2}$ represent the complementary angle of maximum vertical angle between object pairs and distance of KMVN respectively.

\subsection{Overall Objective}

The supervised learning loss for the detection of labelled source samples is denoted as $L_{det}$, which is composed of classification and regression error for RPN and RCNN. Combining detection loss and our introduced losses for multi-view MRD, the final loss function of MMRDN is written as:
\begin{equation}
L = \lambda_1 L_{det} + \lambda_2 L_{vmfml} + \lambda_3 L_{st} + \lambda_4 L_{grl} + \lambda_5 L_{D_{rel}}
\end{equation}
where $L_{D_{rel}}$ is cross entropy loss for manipulation relationship classification.

\begin{table*}[t]
\caption{RECALL AND PRECISION OF VMRD BASED ON REGRAD DATASET}
\begin{center}
\begin{tabular}{|c|c|c|c|c|c|c|c|c|c|}
\hline
\textbf{Metric}&\multicolumn{9}{c|}{\textbf{Recall}} \\
\cline{2-10} 
\textbf{Perspctives} & \multicolumn{3}{c|}{\textbf{\textit{seen}}}& \multicolumn{3}{c|}{\textbf{\textit{unseen}}}& \multicolumn{3}{c|}{\textbf{\textit{real-world}}} \\
\hline
  & parent& child& no-rel& parent& child& no-rel& parent& child& no-rel  \\
 \hline
VMRN & 16.47& 8.23& 96.08& 4.34& 3.69& 98.60& 17.5& 5.00& 97.21  \\
\hline
GGNN+VMRN & 21.30& 14.13& 95.00& 12.17& 10.22& \textbf{98.68}& 17.50& 12.50& 96.99  \\
\hline
\cite{sr} &28.57& 25.00& 95.03& 37.50& 28.57& 95.03& 12.25& 11.36& \textbf{98.35}  \\
\hline
\bf only CPM & 38.98& 38.82& 95.84& 37.71& 37.70& 95.77& 15.46& 15.20& 84.42  \\
\hline
\bf CPM+VMFML & 41.84& 43.90& 95.99& 40.22& 40.32& 96.12& 18.89& 17.31& 82.28  \\
\hline
\bf ours & \textbf{43.48}& \textbf{44.93}& \textbf{96.08}& \textbf{46.33}& \textbf{41.74}& 96.02& \textbf{22.76}& \textbf{19.29}& 81.80  \\
\hline
\textbf{Metric}&\multicolumn{9}{c|}{\textbf{Precision}} \\
\cline{2-10} 
\textbf{Perspctives} & \multicolumn{3}{c|}{\textbf{\textit{seen}}}& \multicolumn{3}{c|}{\textbf{\textit{unseen}}}& \multicolumn{3}{c|}{\textbf{\textit{real-world}}} \\
\hline
  & parent& child& no-rel& parent& child& no-rel& parent& child& no-rel  \\
 \hline
VMRN & 11.43& 8.25& 97.50& 10.05& 11.64& 95.47& 17.83& 7.92& 90.92  \\
\hline
GGNN+VMRN & 12.81& 10.80& 97.49& 20.97& 20.61& 96.34& 20.29& 16.95& 90.52  \\
\hline
\cite{sr} & 0.44& 0.44& \textbf{99.94}& 0.66& 0.44& \textbf{99.94}& 20.56& 19.91& \textbf{92.41}  \\
\hline
\bf only CPM & 21.56& 12.85& 98.64& 19.43& 12.88& 98.77& \textbf{52.60}& \textbf{49.35}& 14.94  \\
\hline
\bf CPM+VMFML & 21.79& 19.61& 98.59& 21.35& \textbf{20.27}& 98.26& 37.66& 37.66& 50.69  \\
\hline
\bf ours & \textbf{21.79}& \textbf{22.22}& 98.61& \textbf{22.05}& 19.87& 98.66& 39.61& 38.96& 56.32  \\
\hline
\end{tabular}
\label{tab1}
\end{center}
\end{table*}

\begin{table*}[t]
\caption{RECALL OF MRD BASED ON DIFFICULTY}
\begin{center}
\begin{tabular}{|c|c|c|c|c|c|c|c|c|c|}
\hline
\textbf{Metric}&\multicolumn{9}{c|}{\textbf{Recall}} \\
\cline{2-10} 
\textbf{Perspctives} & \multicolumn{3}{c|}{\textbf{\textit{simple}}}& \multicolumn{3}{c|}{\textbf{\textit{middle}}}& \multicolumn{3}{c|}{\textbf{\textit{hard}}}\\
\hline
  & parent& child& no-rel& parent& child& no-rel& parent& child& no-rel \\
 \hline
VMRN & 22.89& 13.25& 95.94& 24.8& 13.2& 95.43 & 14.79& 7.03& 96.02  \\
\hline
GGNN+VMRN & 30.24& 20.15& \textbf{97.08}& 13.20& 10.01& \textbf{98.68}& 12.50& 11.548& \textbf{97.98} \\
\hline
\cite{sr} &\textbf{47.50}& \textbf{50.82}& 96.05& 42.49& 39.59& 94.34& 39.61& 41.67& 92.84  \\
\hline
\bf only CPM & 38.76& 43.62& 96.73& 49.80& 49.77& 94.99& 42.51& 41.79& 92.94  \\
\hline
\bf CPM+VMFML & 44.44& 42.48& 96.82& 52.24 & 49.89& 95.03& 42.44& 41.59& 92.88  \\
\hline
\bf ours & 44.14& 47.02& 96.79& \textbf{58.90}& \textbf{57.96}& 95.08& \textbf{50.21}& \textbf{48.11}& 92.79  \\
\hline
\end{tabular}
\label{tab3}
\end{center}
\end{table*}

\section{Experiments}

\subsection{Training Details}

Mask RCNN\cite{maskrcnn} is applied to segment and classify the instance. The learning rate is 0.01, the batch size is 5 and the momentum is 0.9. And $\kappa$ in Eq.\ref{psij} and Eq.\ref{ptij} is set to 20. $\lambda_2$  is set to 10 and all of others are set to 1. The REGNet algorithm\cite{regnet} is applied to auxiliarily demonstrate the effectiveness of MMRDN in real robot experiments.

\subsection{Dataset and Metrics}

\textbf{Dataset} We implement the experiments both on REGRAD dataset\cite{regrad} automatically collected in the virtual environment and a few real-world data.
The marginal distribution of the simulated data and real-world data is different, but the distribution of labels remain the same.
The REGRAD dataset\cite{regrad} has nine camera views data in the ``train'' part of which we use four views, and we evaluate our method on ``seen val'' part. The camera views of training and validate process are shown in Fig.\ref{Network}


\textbf{Metrics} 1) Precision and Recall: Similar to most classification tasks, we test the class precision and recall of three classes. \textit{Obj. Rec.} and \textit{Obj. Prec.} proposed by Zhang\cite{vmrn} will be dominated by the number of unrelated object pairs. 2) Scene Accuracy (SA): this metric tests the accuracy based on the whole scene. In this setting, the scene is considered correct only when all possible stacked object pairs are predicted correctly. We evaluate the performance on scenes with different numbers of objects to demonstrate the performance on scenes of varying complexity.

\subsection{Main results}

We compare the performance of MRD with previously state-of-the-art algorithms. The header in them with \textbf{\textit{seen}}, \textbf{\textit{unseen}} and \textbf{\textit{real-world}} denote views same with training, views different from training and random views in the real world, which are shown in Fig.\ref{Network}. The header in them with ``parent'', ``child'' and ``no-rel'' denote the ``parent-child'' relationship, the ``child-parent'' relationship and the ``no relationship'' respectively. The inference time of each scene is about 0.865s where the process of point cloud processing is about 0.682s.

\begin{figure}[htbp]
\centering
\includegraphics[width=0.45\textwidth]{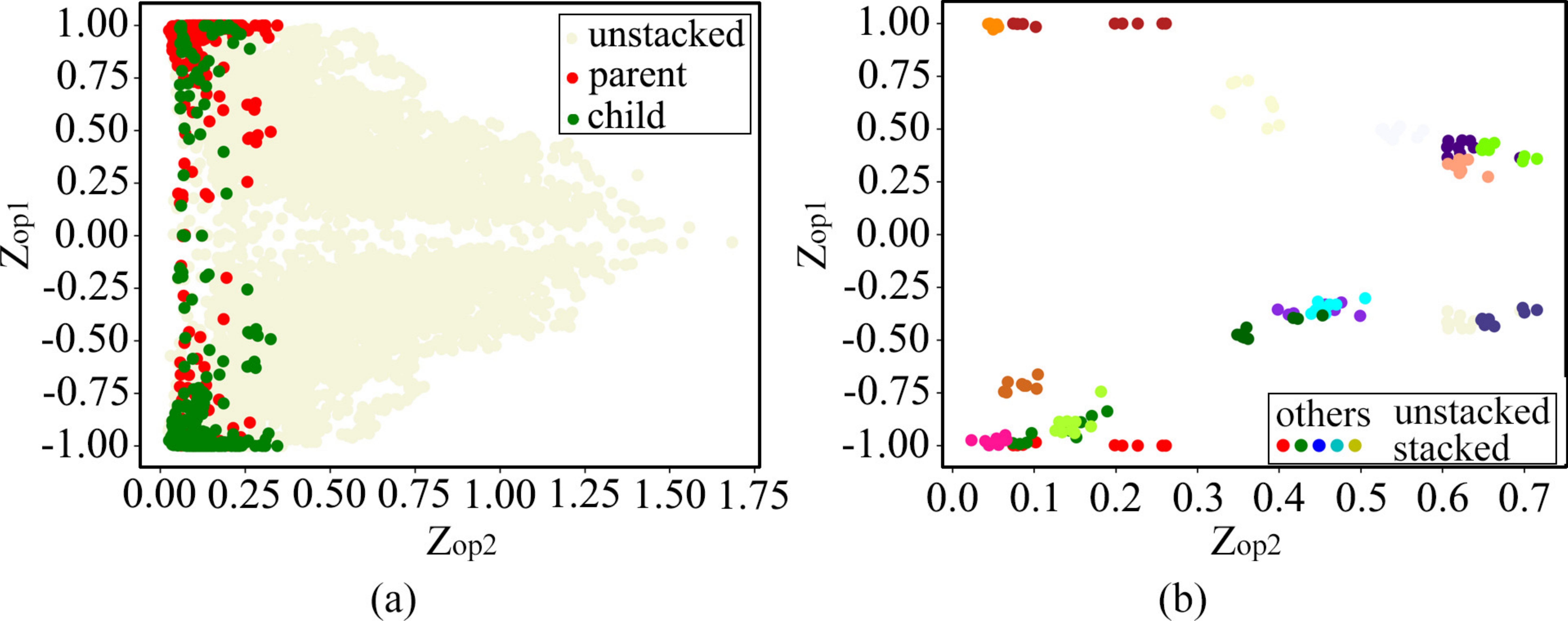}
\label{p-c}
\caption{The distribution of representation from KMVN for MRD. (a) The distribution of $Z_{op1}$ and $Z_{op2}$ of different relationships. (b) The distribution of $Z_{op1}$ and $Z_{op2}$ for different object pairs under different views, where different colors denote different object pairs, and different points of the same color represent data from different views.}
\label{distribution}
\end{figure}

\textbf{Results on different views} Table.\ref{tab1} have shown the Recall and Precision on MRD under different views. 
The results show that the performance of MRD have been improved under both seen views data and unseen views data and even in real-world data. 
Fig.\ref{distribution}(c) shows the distribution of $Z_{op1}$ and $Z_{op2}$ of object pairs in a scene from different views, where different colors represent different object pairs, and different points of the same color represent data from different views, indicates that views have little effect on $Z_{op1}$ and $Z_{op2}$ between the same object pair.

\begin{figure}[htbp]
\centering
\includegraphics[width=0.45\textwidth]{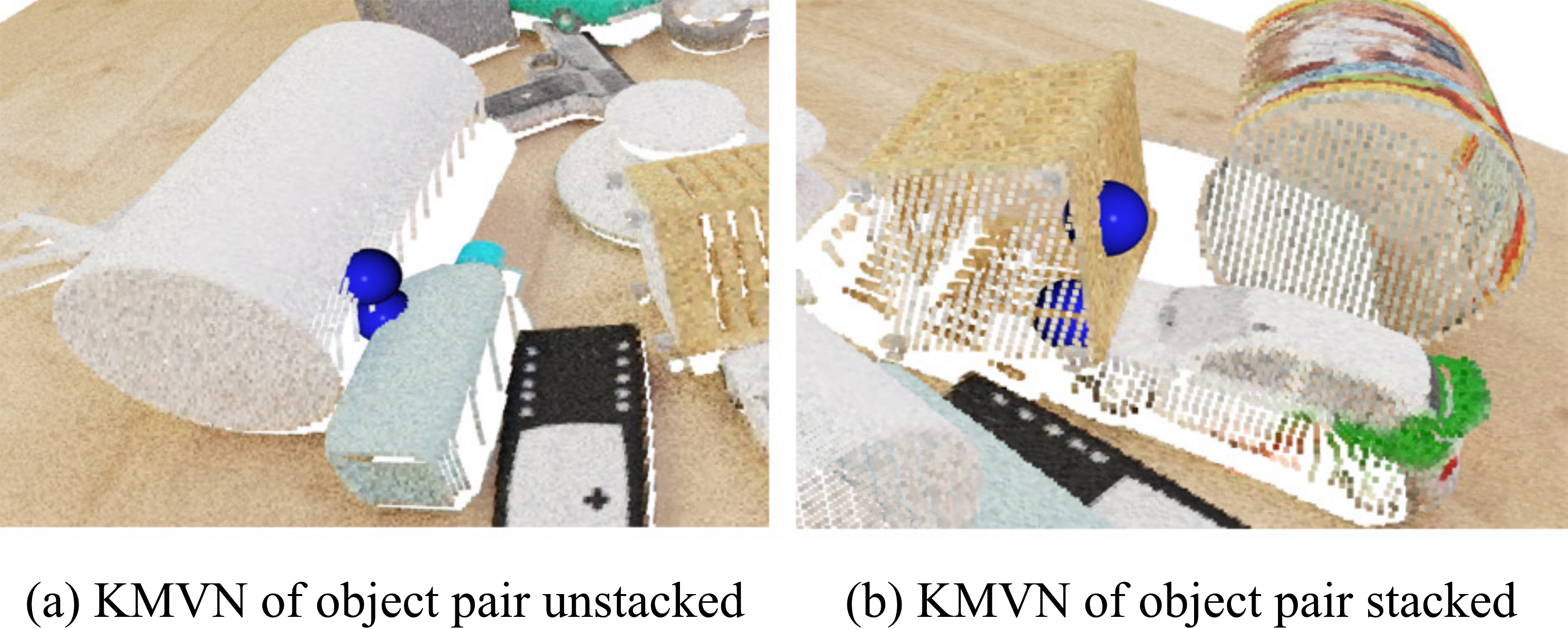}
\label{a}
\caption{Limitation of KMVN. (a) The KMVN of ``can'' and ``bus'' which are unstacked. (b) The KMVN of ``basket'' and ``car'' which are stacked.}
\label{bid-pairs}
\end{figure}

The increase in Recall of the ``parent'' and ``child'' relationship is more pronounced than Precision, since many negative samples (unstacked object pairs) will be predicted as positive samples (stacked object pairs). We explain this phenomenon in Fig.\ref{distribution} where colored dots denote the stacked object pairs and light dots denote the unstacked object pairs.
Fig.\ref{distribution}(a) shows that the distribution of $Z_{op1}$ and $Z_{op2}$ of unstacked object pairs is close to it of stacked object pairs. 
This happens because some objects unstacked to each other are very close together and they have obvious differences in size. For example in Fig.\ref{bid-pairs}(a), the ``can'' is taller and bigger than the ``bus'' and they are close in space but unstacked; and in Fig.\ref{bid-pairs}(b), the ``basket'' and the ``car'' is stacked, but the direction and distance distributions of KMVN between the two object pairs are propinquity.

\begin{figure}[htbp]
  \centering
  \includegraphics[width=.4\textwidth]{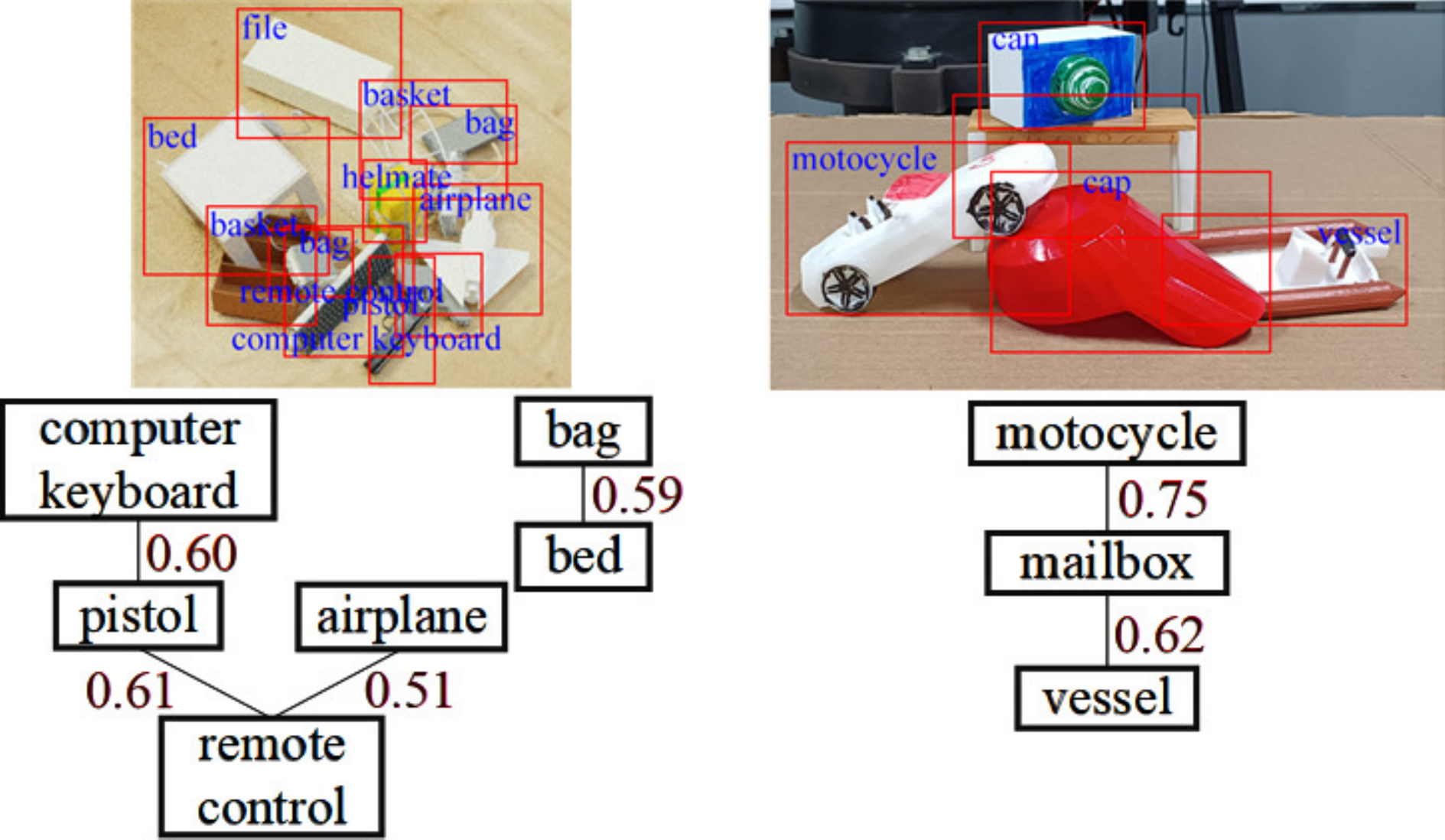}
  \caption{Result examples. Examples for scenes in synthetic data and real-world data from different views and the red numbers denote the scores.}
  \label{relation}
\end{figure}

Besides, we divide the test dataset according to the number of objects. Table.\ref{img-acc} shows the SA of MMRDN in different parts. The results show that SA improves in most scenarios and the mean in all scenarios is improved by 5.6\% compared to the previous state-of-the-art algorithm. We show some detection examples in Fig.\ref{relation} 

\textbf{Results on different degree of occlusion} Table.\ref{tab3} shows the Recall on different degrees of occlusion. We define the simple scenes where the number of the stacked object pairs is less than or equal to 5, middle scenes where the number of the stacked object pairs is in $[5,10]$, and hard scenes where the number of objects is more than $10$. First three lines of the Table.\ref{tab3} are previous algorithms, and last three rows are our method.
The results show that our algorithm performs better in complex scenes with more stacked objects. 

\begin{table}[h]
\caption{IMAGE-WISE TRIPLET ACCURACY OF VMRD BASED ON DATESET REGRAD}
\label{img-acc}
\begin{center}
\begin{tabularx}{.45\textwidth}{>{\centering}m{0.05\textwidth}>{\centering}m{0.05\textwidth}>{\centering}m{0.05\textwidth}>{\centering}m{0.05\textwidth}>{\centering}m{0.05\textwidth}c}
	\hline
    \bf Alog. & Total & Two & Three & Four & Five \\
    \hline
     VMRN\cite{vmrn} & 15.30 & 1.00 & 1.00 & 55.38& 43.75  \\
     GGNN\cite{ggnn-vmrn} & 19.20 & 1.00 & 1.00 & 63.08& 58.04  \\
     \cite{sr} & 20.60 & 1.00 & 1.00 & 67.69 & 57.14  \\
    \bf ours & \textbf{26.20} & \textbf{1.00} & \textbf{1.00} & \textbf{70.77}& \textbf{61.61}  \\
    \hline
    \bf Alog. & Six & Seven & Eight & Nine & Ten  \\
    \hline
     VMRN\cite{vmrn} & 25.21 & 15.19 & 11.67 & 1.10& 0.0  \\
     GGNN\cite{ggnn-vmrn} & 32.77 & 22.78 & 15.83 & 1.10 & 0.0  \\
     \cite{sr} &  33.61 & 31.65 & 13.33 & 1.10 & 0.0  \\
    \bf ours & \textbf{45.38} & \textbf{37.97} & \textbf{20.83} & \textbf{6.59} & \textbf{6.78}  \\
    \hline
    \bf Alog. & Eleven & Twelve & Thirteen & Fourteen & Fifteen  \\
    \hline
     VMRN\cite{vmrn} & 0.0 & 4.92 & 0.0 & 0.0& 0.0  \\
     GGNN\cite{ggnn-vmrn} & 0.0 & 6.56 & 1.39 & 0.0 & 0.0  \\
     \cite{sr} & 0.0 & 6.56 & \textbf{13.89} & 7.69 & 0.0  \\
    \bf ours & 0.0 & \textbf{13.11} & 6.94 & \textbf{12.82}& 0.0  \\
    \hline
\end{tabularx}
\end{center}
\end{table}

\textbf{Ablation study} We show the results of the ablation experiments in the last three rows of Table.\ref{tab1}, each of which denotes the contact point module only (CPM only), contact point module and instance-level alignment module in instance feature among source domains with no target data (CPM+VMFML), and the total MMRDN (ours), to demonstrate the superiority of each part of the MMRDN. The results show that the representations from KMVN play a key role and the instance feature is a blessing.

\begin{figure}[htbp]
  \centering
  \includegraphics[width=.47\textwidth]{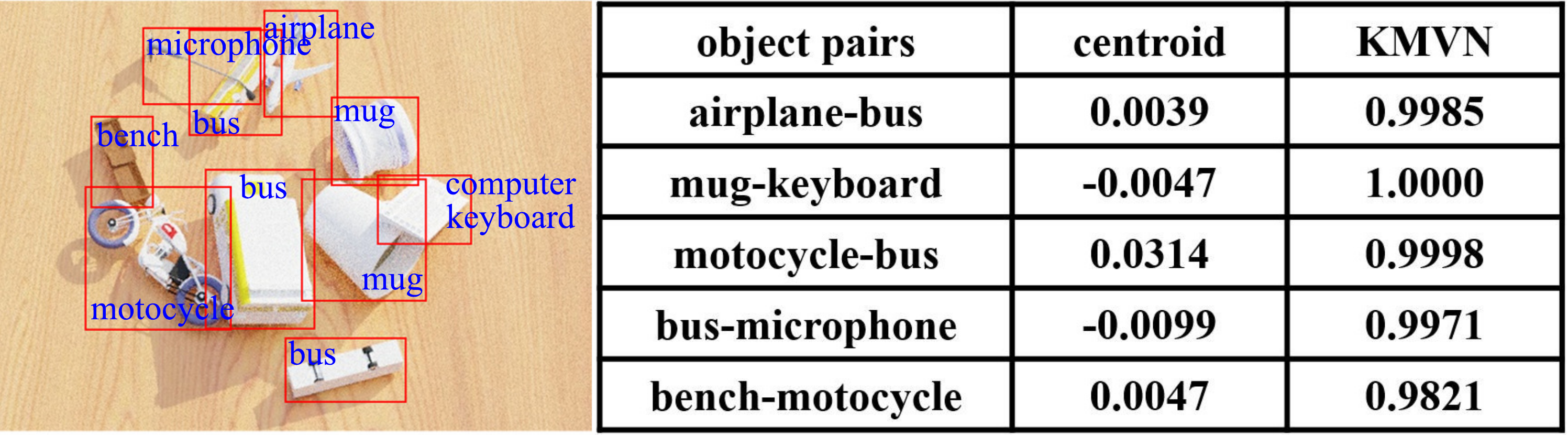}
  \caption{Comparison between KMVN and centroid. The values in the second column denote the difference in z between the centroid coordinates of the two objects, and the third column denotes $Z_{op1}$.}
  \label{op}
\end{figure}

\textbf{Comparison between KMVN and another special point} 
The centroid is a kind of special point on the object. Many researches using 3D point clouds will focus on the centroid, while we select the KMVN. Because in some stacked scenes, the relative positions of the centroids can not correctly represent the relative positions between stacked objects pairs. Although object pairs are stacking, the coordinates of the centroid of the object above are not necessarily higher than those of the object below. For example, the scene is shown in Fig.\ref{op}, where ``computer keyboard'' is supported by ``mug'', but the Z coordinate of ``computer keyboard'' is lower than it of ``mug''.

\begin{figure}[h]
  \centering
  \includegraphics[width=.43\textwidth]{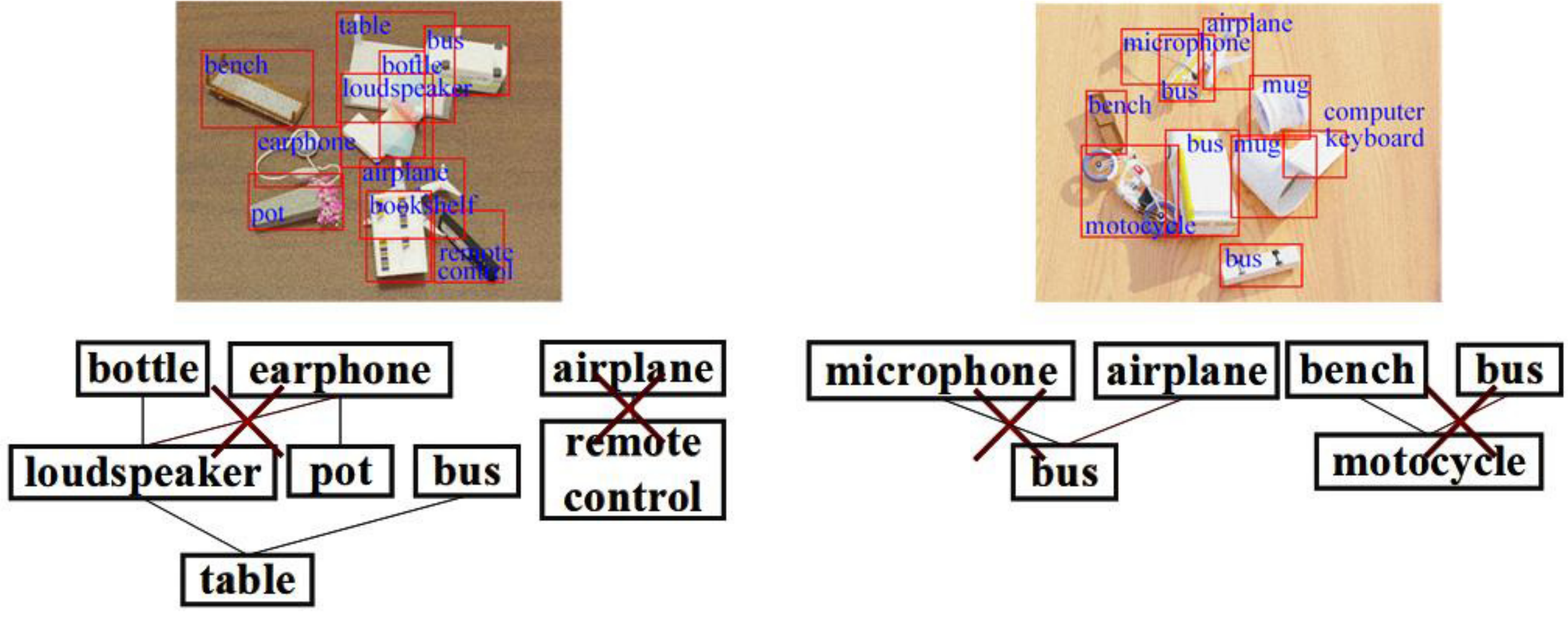}
  \caption{Example scenes that has wrong labels. In the left image, there are redundant relationships. In the right image, they reverse the relationships between object pairs within ``parent-child'' and ``child-parent''.}
  \label{wrong label data}
\end{figure}

\textbf{Label error correction} In theory, when an object pair is ``child-parent'' relationship, $Z_{op1} > 0$ should be satisfied.
However, the result from Fig.\ref{distribution}{b} shows that there are some object pairs whose $Z_{op1} < 0$ where green dots and red dots denote the ``child-parent'' and ``parent-child'' relationship respectively. Therefore, we check some scenes based on $Z_{op1}$ and find a few relationship labels of some object pairs are wrong. We show some mislabeled data in Fig.\ref{wrong label data}.

\section{CONCLUSIONS}


Multi-view MRD suffers from the domain shift due to occlusion difference in different views. In this paper, we propose a novel multi-view fusion framework to learn the consistent representations among multiple views for MRD in object stacked scenes. 
Our approach models the relative relationship of object pairs by KMVN and align instance feature from different views by VMF distribution. Experiments are conducted to evaluate our approach in scenarios from multiple views. The experimental results show that our approach outperforms the previous methods and achieves the state-of-the-art performance on multi-view MRD.


\addtolength{\textheight}{-9cm}



\newpage
\bibliographystyle{IEEEtran}
\IEEEtriggeratref{11}
\bibliography{root}

\end{document}